\documentclass[conference]{IEEEtran}

\IEEEoverridecommandlockouts

\usepackage{cite}
\usepackage{todonotes}
\usepackage{flushend}
\usepackage{amsmath,amssymb,amsfonts}
\usepackage{algorithmic}
\usepackage{float}
\usepackage{graphicx}
\usepackage{enumitem}
\usepackage{caption}
\usepackage{subcaption}
\usepackage{textcomp}
\usepackage{xcolor}
\usepackage{hyperref}
\usepackage{booktabs,siunitx,tabularx, stfloats}
\usepackage{threeparttable}
\usepackage{svg}
\usepackage{diagbox}
\usepackage{fancyhdr}
\usepackage{pgf}
\usepackage{bbding}
\newcommand{\orcid}[1]{\href{https://orcid.org/#1}{\includegraphics[width=6pt]{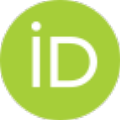}}}
\renewcommand{\baselinestretch}{0.905}

\def\BibTeX{{\rm B\kern-.05em{\sc i\kern-.025em b}\kern-.08em
    T\kern-.1667em\lower.8ex\hbox{E}\kern-.125emX}}
\newcommand{\printfnsymbol}[1]{%
  \textsuperscript{\@fnsymbol{#1}}%
}

\begin{document}

\title{Explainable Artifacts for\\ Synthetic Western Blot Source Attribution}

\author{João P. Cardenuto\textsuperscript{1,4}\orcid{0000-0002-8370-6329}, Sara Mandelli\textsuperscript{2}\orcid{0000-0003-3811-003X}, Daniel Moreira\textsuperscript{4}\orcid{0000-0001-9757-5756}, Paolo Bestagini\textsuperscript{2}\orcid{0000-0003-0406-0222}, Edward Delp\textsuperscript{3}\orcid{0000-0002-2909-7323}, Anderson Rocha\textsuperscript{1}\orcid{0000-0002-4236-8212}\\ 
\small{\textsuperscript{1}Artificial Intelligence Lab., \hyperlink{recod.ai}{Recod.ai}, Institute of Computing, Universidade Estadual de Campinas (UNICAMP), Campinas, SP, Brazil.} \\
\small{\textsuperscript{2}Dipartimento di Elettronica, Informazione e Bioingegneria, Politecnico di Milano, 20133 Milan, Italy.} \\
\small{\textsuperscript{3}Purdue University, School of Electrical and Computer Engineering, West Lafayette, IN, USA} \\
\small{\textsuperscript{4}Department of Computer Science, Loyola University Chicago, Chicago, IL, USA.} \\
Email: phillipe.cardenuto@ic.unicamp.br\\
\thanks{This research was funded by the São Paulo Research Foundation (Fapesp) under the grants Horus \#2023/12865-8, and \#2020/02211-2.
It was also sponsored by the Defense Advanced Research Projects Agency (DARPA), the Department
of Health and Human Services (HHS), and the Air Force Research Laboratory (AFRL) under AFRL agreement FA8750-16-
2-0173.}}

\maketitle

\fancypagestyle{firstpage}{
    \renewcommand{\headrulewidth}{0pt}%
    \fancyhf{}
    \fancyfoot[L]{\footnotesize\textcopyright 2024 IEEE.  Personal use of this material is permitted. Permission from IEEE must be obtained for all other uses in any current or future media, including reprinting/republishing this material for advertising or promotional purposes, creating new collective works, for resale or redistribution to servers or lists, or reuse of any copyrighted component of this work in other works.}
 }
 
\thispagestyle{firstpage}

\begin{abstract}
Recent advancements in artificial intelligence have enabled generative models to produce synthetic scientific images that are indistinguishable from pristine ones, posing a challenge even for expert scientists habituated to working with such content. When exploited by organizations known as paper mills, which systematically generate fraudulent articles, these technologies can significantly contribute to the spread of misinformation about ungrounded science, potentially undermining trust in scientific research. While previous studies have explored black-box solutions, such as Convolutional Neural Networks, for identifying synthetic content, only some have addressed the challenge of generalizing across different models and providing insight into the artifacts in synthetic images that inform the detection process. This study aims to identify explainable artifacts generated by state-of-the-art generative models (e.g., Generative Adversarial Networks and Diffusion Models) and leverage them for open-set identification and source attribution (i.e., pointing to the model that created the image).
\end{abstract}


\section{Introduction}
% What is the background problem?
% Why is it important?
% How we address the problem?
Numerous problematic scientific articles have recently been reported, presenting distinctive features that suggest they were systematically produced. Dr.~J.~Christopher, an editor of FEBS PRESS, was the first to report multiple manuscripts with doctored figures received by their editorial board~\cite{Christopher2018}. These figures shared similar backgrounds and unique characteristics despite being attributed to different authors.
Following this report, thousands of other articles have been flagged as systematically produced and subsequently retracted by other journals~\cite{Else2021}, with most published in the medical and biological fields~\cite{Sanderson2024}. 

The case has been attributed to potentially illegal organizations, known as paper mills, which provide scientific writing and publishing services for papers seemingly lacking scientific merit~\cite{Byrne2020}. Recent investigations have found that this industry generates millions of dollars worldwide and has been ``bribing editors and planting their agents on editorial boards to ensure publication''~\cite{Wise2024}.

To worsen the issue, recent advancements in generative Artificial Intelligence (AI) models, along with their increasing accessibility, could aid paper mills in expanding their production.
Previous work has demonstrated that AI synthetic scientific images can be indistinguishable from genuine ones~\cite{Qi2020} and could threaten scientific integrity. Such a threat may have already materialized, as recent publications have been retracted for using AI-generated images~\cite{Guo2024}.

Western blots are the most concerned type of scientific image in this context because they are easily generated using AI~\cite{Qi2020} and are frequently used in publications by paper mills~\cite{Christopher2018, Byrne2020}. These images derive from biomedical experimental procedures used in laboratories to detect and measure protein levels.
More than 400,000 research works rely on these images, accounting only for those listed in PubMed~\cite{wblots2023}, a large repository of biomedical articles.

A possible approach to identify paper mills involves analyzing the similarity of systematically produced images with image provenance analysis~\cite{Cardenuto2023}. However, generative AI could easily produce never-before-seen images, which makes their identification more challenging.

Another method is to identify the source model that generated the AI figures, tracking mills by the models they are using. When designing such a solution, we should consider the complex nature of paper mills and the severe implications of falsely accusing authors. A solution to this problem should not only determine the source model of an image but also provide a clear explanation for its decision. 

Therefore, in this work, we explore forensic solutions to identify and attribute the source of synthetic Western blots. We rely on explainable low-level artifacts from AI generation methods.
Our contributions are threefold:
\begin{enumerate}
    \item An analysis of low-level artifacts present on synthetic Western blot images;
    \item New methods to expose AI artifacts, namely (i) by analyzing image patches using the Fourier spectrum and 
    (ii) by the analysis of texture-based features;
    \item An analysis of the residual-noise impact on exposing synthetic artifacts.
\end{enumerate}

The code and dataset from our research are available at \url{https://github.com/phillipecardenuto/ai-wblots-detector}

\section{AI Generation Artifacts}
\label{sec:ai_artifacts}

This section investigates possible sources of synthetic generation artifacts and lists some promising features that can be exploited to spot AI-generated images and perform generation model source attribution.

\subsection{Common AI Generation Artifacts}

\begin{figure}[t]
    \centering
    \begin{subfigure}[t]{0.23\textwidth}
        \centering
        \includegraphics[width=\textwidth]{ 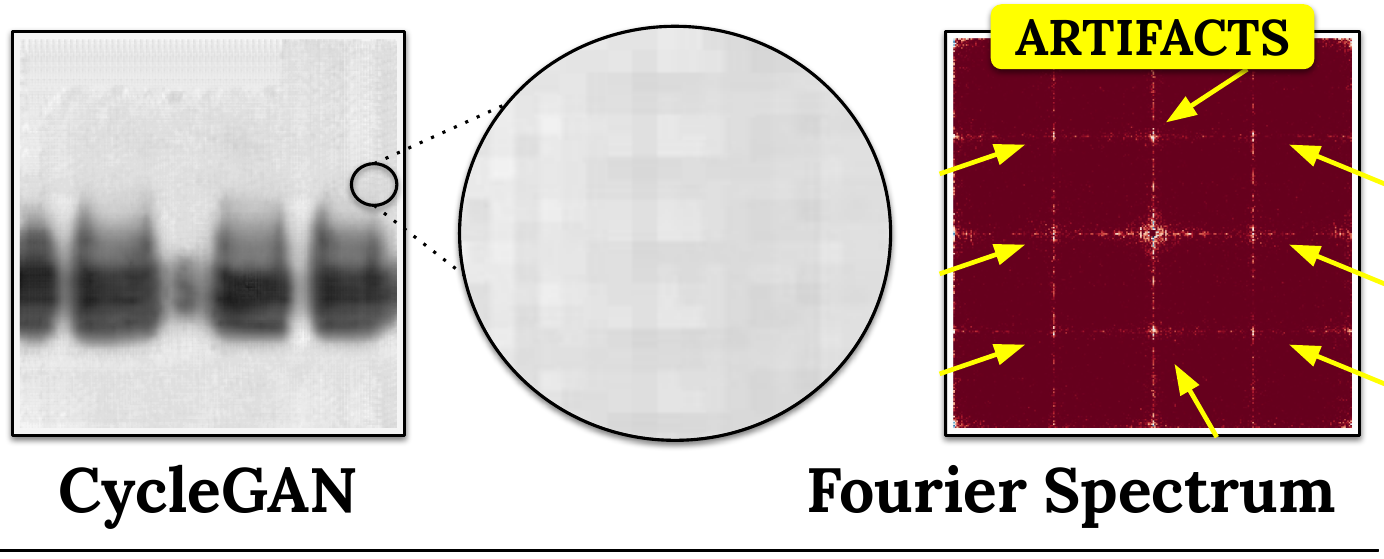}
        \caption{Checkerboard}
    \end{subfigure}%
    ~ 
    \begin{subfigure}[t]{0.23\textwidth}
        \centering
        \includegraphics[width=\textwidth]{ 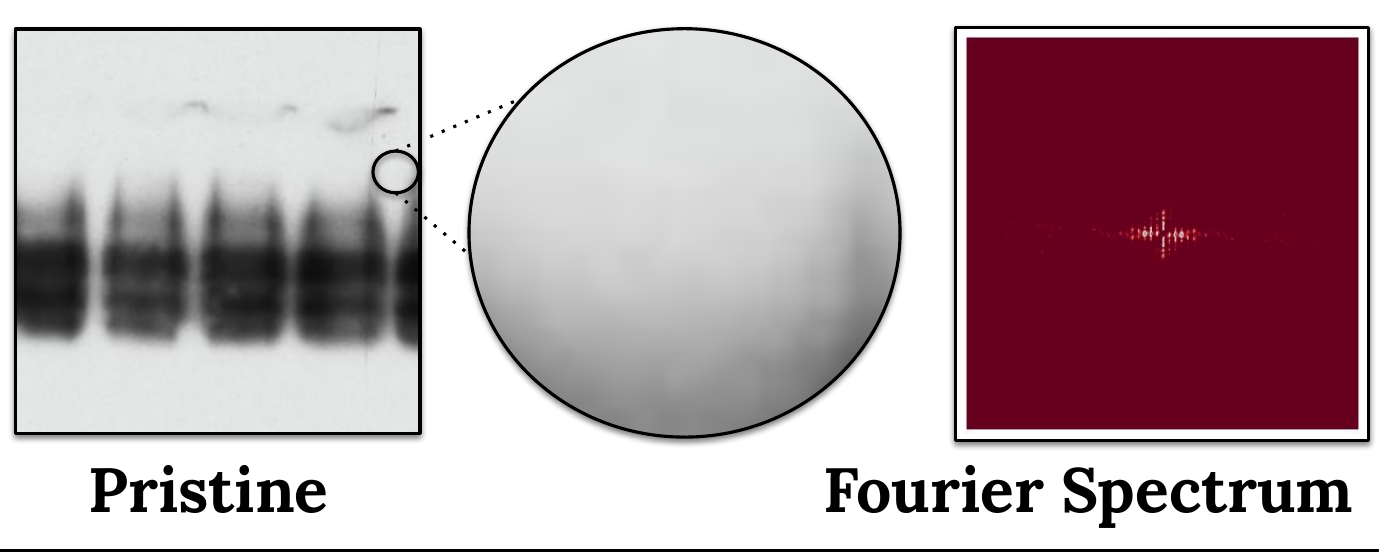}
        \caption{No checkerboard}
    \end{subfigure}
    \caption{Comparison between a CycleGAN~(a) and a pristine~(b) Western blot image. The CycleGAN image contains checkerboard artifacts visible when zooming into the image. The highlighted Fourier spectrum peaks (see the yellow arrows) also indicate the presence of those artifacts.}
    \label{fig:checkerboard-art}
\vspace{-0.6cm}
\end{figure}

Most generative AI models work as autoencoders~\cite{CardenutoAI2023}. They first encode an input signal into a latent space and then decode it into an intelligible media such as text, images, or audio.
In the case of images, the process
transforms a one-dimensional vector into a two-dimensional array (i.e., the output image). During this process, most models perform an upsampling operation to increase the size of the image, which typically adds specific artifacts~\cite{odena2016deconvolution}.
Such sampling artifacts occur unnaturally and at a periodic rate that can be exposed using, for instance, Fourier spectrum analysis~\cite{Gragnaniello2021, corvi2023detection}. Due to their nature, we refer to them as periodic artifacts.

Checkerboard artifacts are well-known examples in this category. As depicts Fig.~\ref{fig:checkerboard-art}, these artifacts appear as repetitive checkerboard-like patterns in AI-generated images. Odena et al.~\cite{odena2016deconvolution} demonstrated that deconvolutional kernels, commonly used by state-of-the-art GAN-based models, are the primary cause of these artifacts. During the deconvolution operation (upsampling), if the kernel size is not divisible by the deconvolution stride, overlapping regions are created in the output for two neighboring pixels from the input image. This overlap occurs at a periodic rate, producing the checkerboard effect. The checkerboard pattern creates distinct high-frequency band peaks in a Fourier spectrum, generating a unique data fingerprint. For instance, popular GAN-based generators like CycleGAN~\cite{CycleGAN2017} and Pix2Pix~\cite{pix2pix2017} use indivisible parameters (kernel size of 3 and stride of 2) for deconvolution.
This results in a noticeable checkerboard effect on their generated samples, as shows Fig.~\ref{fig:checkerboard-art}.

% How to avoid these artifacts?
Odena et al.~\cite{odena2016deconvolution} suggested using a resize-convolution operation to avoid these artifacts. This operation first resizes the signal using nearest-neighbor interpolation and then applies a standard convolutional kernel, avoiding the overlap. This fix has been used on the upsampling implementation of recent generators to improve their output~\cite{ho2020denoising}.
However, while this approach may address the checkerboard pattern, it still produces a linear combination of the pixels' neighborhood in the output image~\cite{odena2016deconvolution}. Therefore, this operation alone cannot eliminate the periodic artifacts, which could still be detected in the Fourier spectrum~\cite{Zhang2019DetectingAS}.

\subsection{Exposing AI generation artifacts}
\label{subsec:artifacts_exposing}
Over the years, the forensic community has proposed multiple solutions to expose periodic artifacts on natural images. 
In the following, we report strategies that are most employed in the state of the art, dividing them into two categories: hand-crafted features and deep learning features. 

\paragraph{Hand-crafted features}
% PMAPs
One of the first works to detect periodic artifacts on natural photos was proposed by Popescu and Farid\cite{Popescu2005}.
They focused on detecting traces of resampling in an image, aiming to understand the specific artifacts that resizing operations leave behind.
The key idea was that resampling creates a correlation among the pixels' neighborhoods, similar to the uneven overlap of checkerboard artifacts. To highlight this correlation, 
they used an Expectation Maximization (EM) algorithm to identify an optimal linear combination that describes how each pixel relates to its neighbors. Using EM, they built a probabilistic map (P-Map) that provides the likelihood of a pixel correlated to its neighbors. This map can identify levels of resampling and possibly expose the periodic artifacts left by the AI generation process.

Following~\cite{Popescu2005}, Kirchner~\cite{Kirchner2008} noticed that the P-Map periodicity originated from the EM optimization's residuum function. He showed that the residuum could be interpreted as a linear filtered version of the original signal that exposes the periodic artifacts.
He thus proposed to replace the EM solution with a specific filter to extract the residuum.

Recent studies have shown that residual noises extracted from AI-generated images can reveal periodic artifacts when analyzing identifiable high-frequency peaks in their Fourier spectrum~\cite{Gragnaniello2021, corvi2023detection}.
Based on that, Bammey~\cite{Bammey2024} computed the Fourier transform of the residual noise and created a feature vector 
from different frequency bands.

He trained a supervised classifier using this feature and achieved promising results in distinguishing fake from pristine images.
In his work, Bammey qualitatively showed that these peaks manifest differently in images generated by different diffusion model classes.

Co-occurrence matrices are another promising approach to exposing synthetic content ~\cite{Nataraj2019, Barni2020}.
In the context of scientific images, Mandelli et al.~\cite{Mandelli2022} employed texture descriptors derived from the gray level co-occurrence matrix (\textit{GLCM}) to distinguish genuine Western blot images from synthetic versions generated through diverse architectures, including GANs and diffusion models.

\paragraph{Deep learning features}
Besides hand-crafted techniques, deep learning-based features have also been used to distinguish pristine images from AI images. 
For instance, Cocchi et al.~\cite{Cocchi2023} used pre-trained Contrastive Language-Image Pretraining (CLIP)~\cite{radford2021learning} and Self-Distillation with No Labels (DINO)~\cite{caron2021emerging} models as feature extractors to distinguish fake from genuine samples. 

Unlike hand-crafted feature extractors, CLIP and DINO are data-driven models trained on large datasets with millions of parameters. This makes it intractable to understand the specific role of each parameter and explain their decisions, hindering the possibility of finding and interpreting their potential biases. Therefore, we should be skeptical about the generalizability of such methods.

\section{Proposed Techniques to Expose AI Artifacts}
\begin{figure}[t]
    \centering
    \includegraphics[width=0.45\textwidth]{ 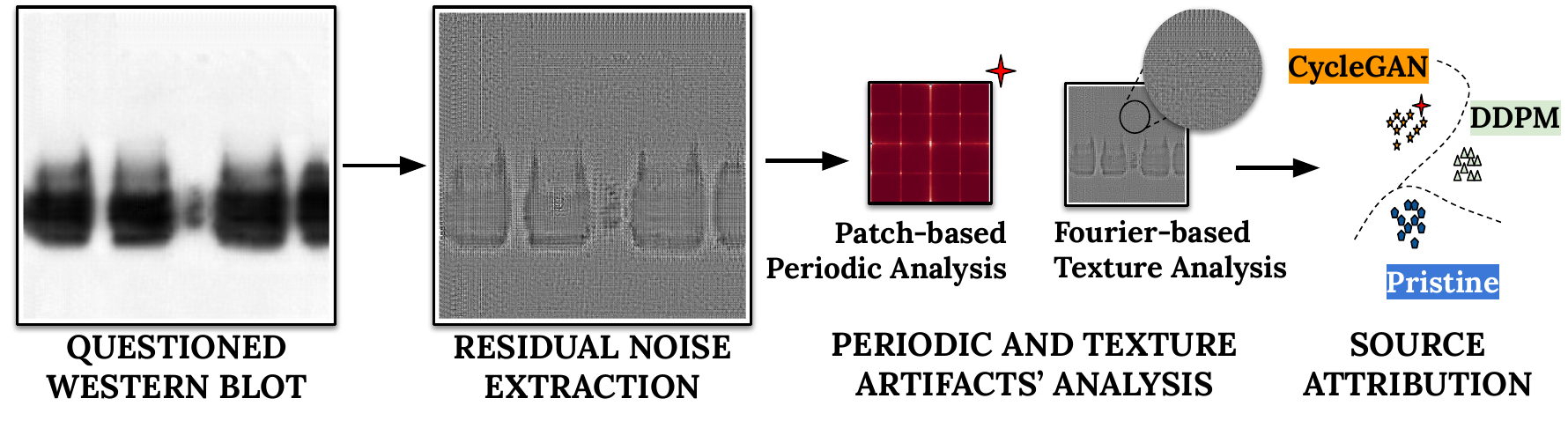}
    \caption{Solution workflow.
    Given a questioned Western blot, we leverage residual noise extraction, periodic artifacts, and texture features' analysis to perform synthetic image detection and AI-generation model source attribution.}
    \label{fig:workflow}
\vspace{-0.6cm}
\end{figure}

Given a questioned Western blot, the problem we address in this research is learning to identify it as either pristine or synthetic and, in the latter case, learning to attribute the AI-generation model that might have been used to create it.
Fig.~\ref{fig:workflow} summarizes the problem and our proposed solution workflow.
As one might observe, we explore residual noise extraction (see Sec.~\ref{sec:exploring-residual-noise}) and the analysis of explainable periodic artifacts (Sec.~\ref{sec:periodic}) and texture features (Sec.~\ref{sec:texture}) to train classifiers that accomplish the task at hand.

Our approach, which involves identifying the distinct artifact-based features of each generative model, has practical implications. It could potentially enable the identification and appropriate action against a paper mill, if necessary.

\subsection{Patch-based Periodic Artifacts}
\label{sec:periodic}

Most low-level artifacts explored for synthetic image detection are exposed after extracting a residual noise from an input image. However, to our knowledge, none of the previously reported studies in Section~\ref{subsec:artifacts_exposing} considered that periodic samples introduced by AI generation should be equally distributed over the image's residual noise, independently of the semantic content depicted in the image. This concept is crucial because analyzing the entire image may raise the possibility of inadvertently including semantic elements from the objects depicted in the image in our analysis.

To minimize the impact of these semantics, we propose to split the image into patches and extract artifacts from the residuals of these patches. We then combine the patch contributions by averaging the Fourier spectrum of their residuals. 

We name this patch-based residual Fourier transform strategy \textit{PATCH-FFT-PEAKS}, while the typical full-image version is \textit{FFT-PEAKS}.

\begin{figure}[t]
    \centering
    \begin{subfigure}[t]{0.24\textwidth}
        \centering
        \includegraphics[width=\textwidth]{ 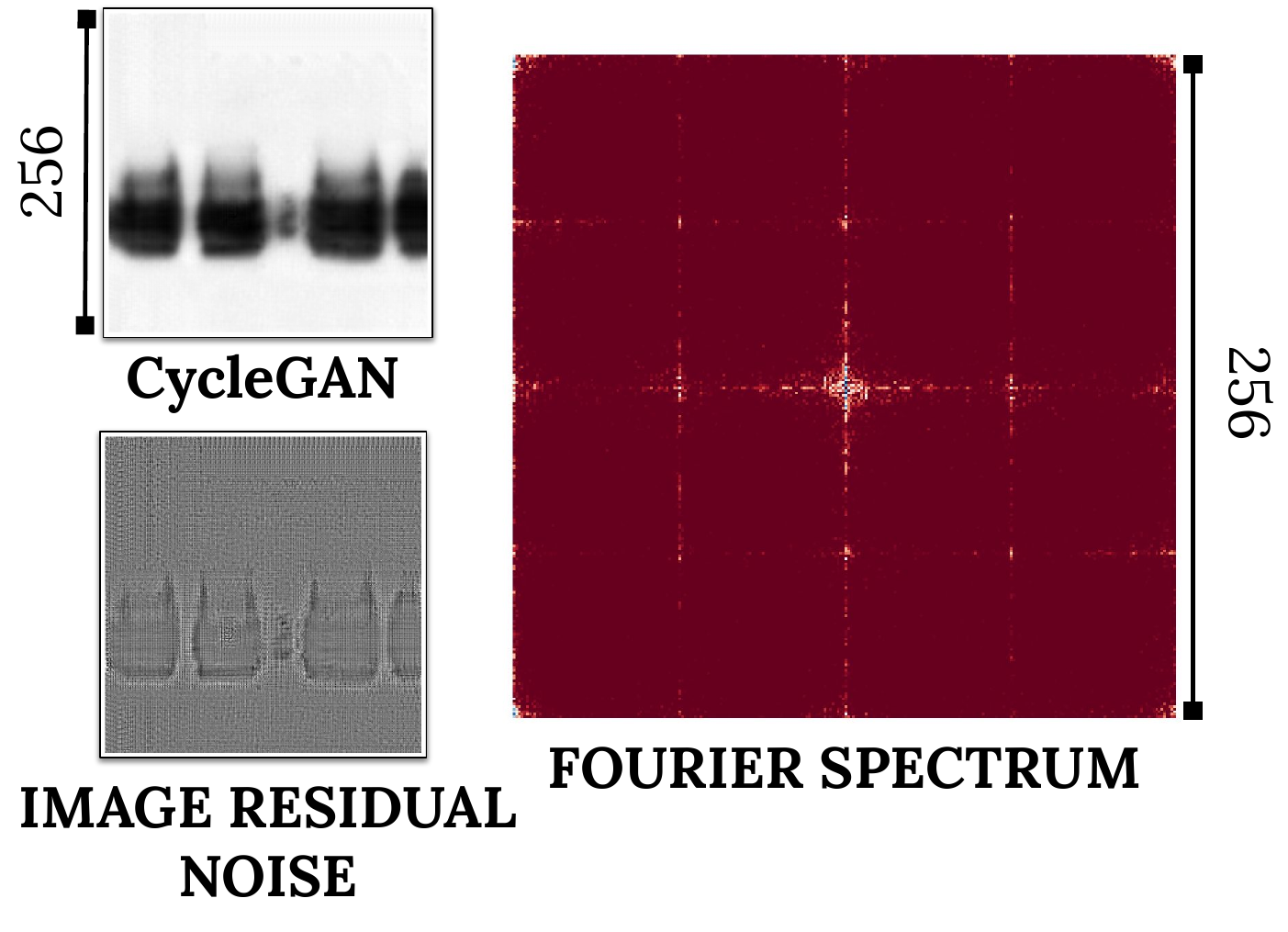}
        \caption{Entire image}
    \end{subfigure}%
    ~ 
    \begin{subfigure}[t]{0.24\textwidth}
        \centering
        \includegraphics[width=\textwidth]{ 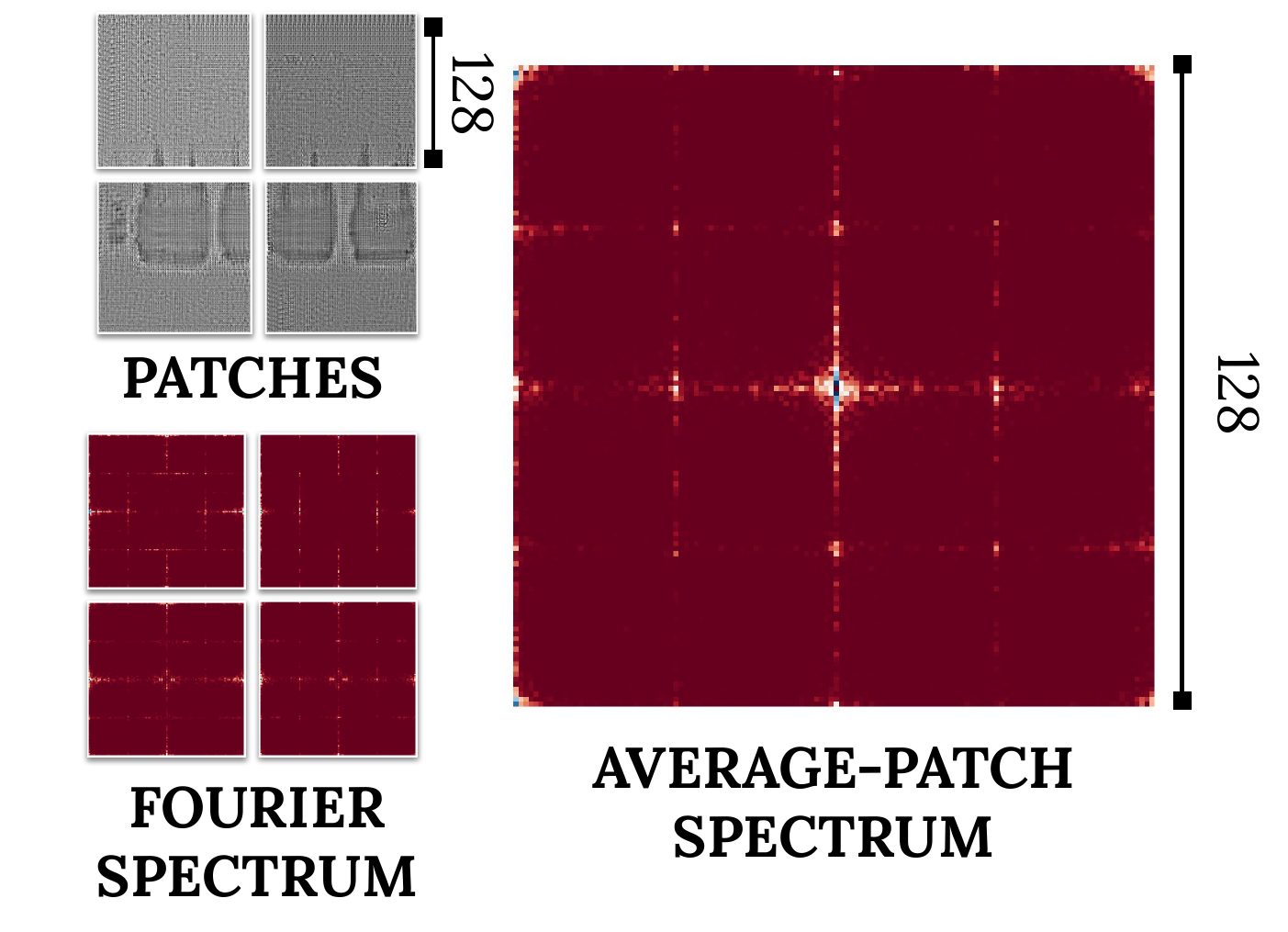}
        \caption{Patch-based}
    \end{subfigure}
    \caption{Comparison between the (a) Fourier calculated over the entire noise residual image (\textit{FFT-PEAKS} strategy) and (b) average-patch Fourier spectrum (\textit{PATCH-FFT-PEAKS} strategy). All spectra are centered in spatial frequencies $(0, 0)$ and are computed over zero-mean signals.}
    \label{fig:combined-spec}
\vspace{-0.5cm}
\end{figure}

Fig.~\ref{fig:combined-spec} illustrates the difference between a spectrum computed directly from the residual noise and one derived from the combined patches. It is worth noticing that the artifacts are more prominently highlighted in the latter case.

%\subsection{Fourier-based Texture Description}
\vspace{-0.08cm}
\subsection{Fourier-based Texture Features}
\label{sec:texture}

Motivated by the successful texture feature extraction in \cite{Mandelli2022}, we propose inspecting the \textit{GLCM} matrices of different generation models. 
Every generation model is associated with a distinct matrix. To highlight this uniqueness, we calculate the Fourier spectrum of each \textit{GLCM} matrix, resulting in a visibly more distinguishable pattern. 
We name this proposed solution as \textit{FFT-GLCM}. 
Similar to the approach in~\cite{Mandelli2022}, during our experiments, we extract contrast-weighted, homogeneity-weighted, dissimilarity-weighted, energy, and correlation-weighted features from the \textit{FFT-GLCM} at distances of 4, 8, 16, and 32, in both horizontal and vertical directions. These features are then concatenated to form a feature vector with 40 dimensions.

\begin{figure}[t]
    \centering
    \includegraphics[width=0.37\textwidth]{ 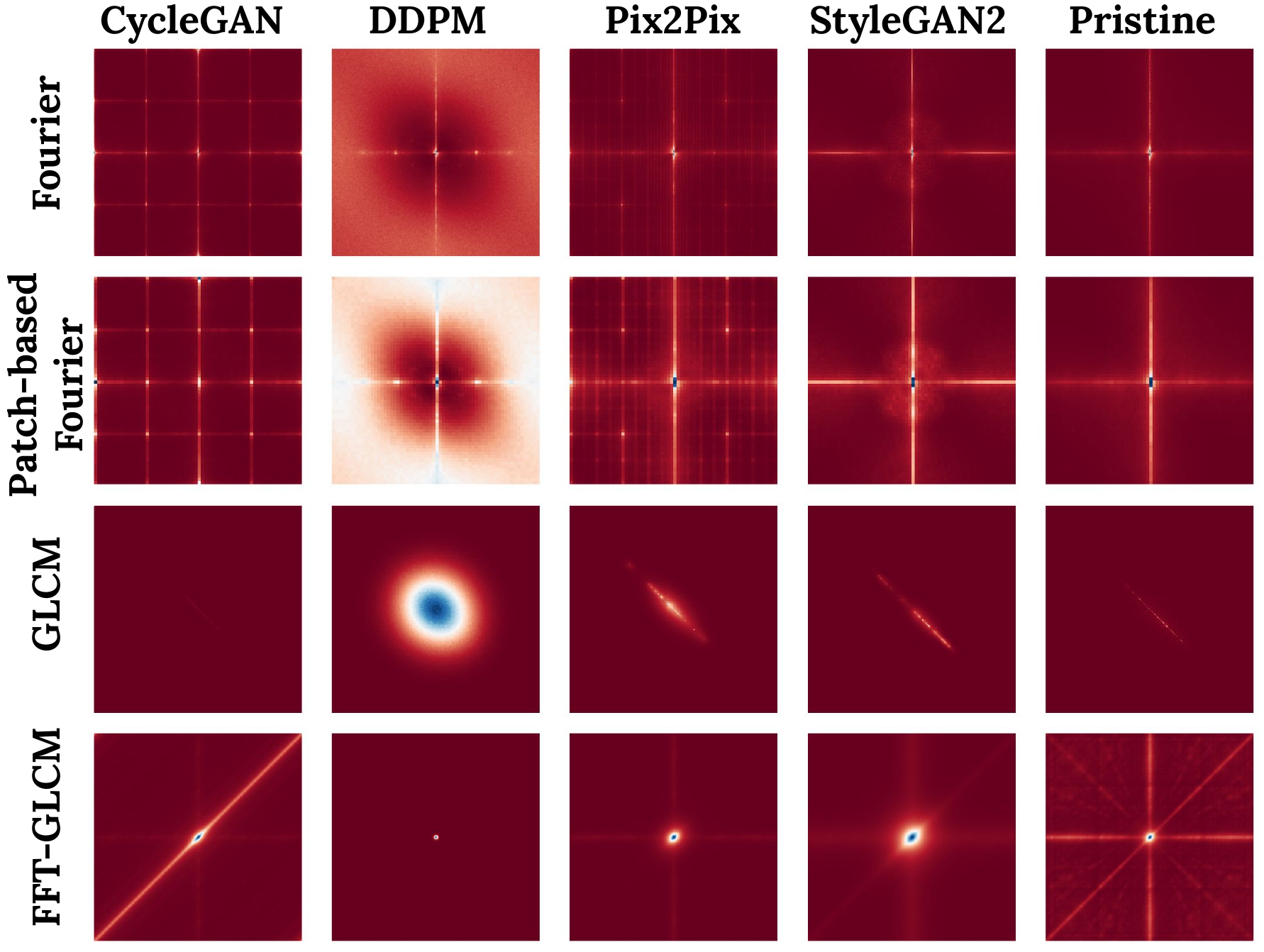} 
    \caption{Different features extracted to expose AI generation artifacts. Each visualization results from an average of $100$ images.
    All spectra are centered in spatial frequencies $(0, 0)$ and are computed over zero-mean signals.
    The Fourier spectra on the same row are depicted over the same scale to help visual comparison. Rows show the explored telltale; columns show different generative AI models and a pristine source.
    }
    \label{fig:comparing-artifacts}
\vspace{-0.5cm}
\end{figure}

Fig.~\ref{fig:comparing-artifacts} compares the Fourier Spectrum (first row), the Patch-based Fourier Spectrum (second row), the \textit{GLCM} (third row), and the proposed \textit{FFT-GLCM} (fourth row) for the average of a hundred samples from each column-wise generation source.
While the Fourier and \textit{GLCM} spectra exhibit faint artifact peaks, patch-based Fourier and \textit{FFT-GLCM} emphasize each generator's artifacts, resulting in unique strong patterns for each generator.
\vspace{-0.08cm}

\subsection{Residual Noise Extraction}
\label{sec:exploring-residual-noise}
The extraction of noise residues from an image to perform forensic investigations is well known and widely used in the forensic community~\cite{Popescu2005, Kirchner2008, Gragnaniello2021}.
Moreover, the state of the art has proposed many different types of residual noise extraction. 
To our knowledge, no study has performed experiments to test how different noise extraction techniques impact a detector's performance. 
This section explores different noise extraction methods to check their impact on the detection of AI generation artifacts.

The selected noise extraction methods are:
\begin{enumerate}
    \item \textit{Mandelli et al.}~\cite{Mandelli2022}: this method convolves a kernel \( T \) (see Eq.~\ref{eq:kernels}) with the image, then computes the difference between the image and the 2D convolution, acting as a high-pass filter.
    \item \textit{Bammey}~\cite{Bammey2024}: similar to \cite{Mandelli2022}, but using the cross kernel \( C \) (see Eq.~\ref{eq:kernels}).
    \item \textit{Gaussian}: similar to \cite{Mandelli2022}, but using a Gaussian kernel with \(\sigma=1\) and radius \( =4\).
    \item \textit{Mean}: similar to \cite{Mandelli2022}, but using a neighborhood mean kernel \( M \) (see Eq.~\ref{eq:kernels}).
    \item \textit{Kirchner}~\cite{Kirchner2008}: it uses the kernel \( K \) (see Eq.~\ref{eq:kernels}) proposed in~\cite{Kirchner2008} to extract the residual noise from an image. 
    \item \textit{P-Map}: instead of noise extraction, this method uses the P-Map solution proposed in~\cite{Popescu2005}
    \item \textit{Non-Local Means}~\cite{Nlmeans2005}:
    It uses the non-local means technique, which preserves texture and periodic elements while denoising the target image~\cite{Nlmeans2005}.
\end{enumerate}
%\vspace{-0.4cm}

The exploited kernels $T, C, M$ and $K$ are listed as follows:
\begin{equation}
\small{
    \begin{split}
        T = \frac{1}{4} \begin{bmatrix}
0 & 1 & 0 \\
1 & 0 & 1 \\
0 & 1 & 0
\end{bmatrix}, \quad &C = \begin{bmatrix}
1 & -1 \\
-1 & 1
\end{bmatrix}, \\
M = \frac{1}{8} \begin{bmatrix}
1 & 1 & 1 \\
1 & 0 & 1 \\
1 & 1 & 1
\end{bmatrix}, \quad      &K = \begin{bmatrix}
-0.25 & 0.50 & -0.25 \\
0.50 & 0 & 0.50 \\
-0.25 & 0.50 & -0.25
\end{bmatrix}.
    \end{split}
\label{eq:kernels}
}
\end{equation}

\section{Experiments and Analysis}

In this section, we present the achieved results by testing the proposed techniques to expose AI artifacts in three different scenarios: (i) closed-set source attribution,  (ii) open-set attribution, and (iii) one-vs-rest source attribution. More details follow in the next lines.

\subsection{Experimental Setup}
We adopt as baselines the data-driven features from DINOv1, DINOv2, and CLIP (Cocchi et al.\cite{Cocchi2023}), texture features (Mandelli et al.~\cite{Mandelli2022}), and periodic artifacts (Bammey's Synthbuster~\cite{Bammey2024}). Additionally, we report results for the extended and proposed texture features (\textit{GLCM} and \textit{FFT-GLCM}) and periodic artifacts (\textit{FFT-PEAKS} and \textit{PATCH-FFT-PEAKS}).
For~\cite{Mandelli2022} and~\cite{Bammey2024} methods, we adopt their original noise extractors. For the other methods, we present the results based on the most effective residual noise extraction technique, defined as Best Noise Extraction (BNE).

\paragraph{\textbf{Closed-set Attribution}}
Here, we approach model attribution as a supervised closed-set classification task. 
The goal is to determine whether the features under investigation can identify a known source model. Given an image, we aim at inferring if it is genuine or generated by specific AI models, all of which were known during the training phase. 
This situation can occur when analysts are already familiar with paper mills' most commonly used generators. Despite its simplicity, this controlled setting serves as a starting point for our investigations.

For this task, we employ well-known, explainable, high-performance machine learning classifiers using the investigated features. Specifically, we use Random Forest (RF) and eXtreme Gradient Boosting (XGBoost).

During the evaluation, 
we use the dataset proposed in~\cite{Mandelli2022}, which consists of synthetic Western blots generated by CycleGAN, Pix2Pix, StyleGan2, and Denoising Diffusion Probabilistic Models (DDPM), as well as a set of pristine Western blot images, totaling five different sources. 
We selected 6,000 images from each source.
We test the models using a cross-validation setup, where half of the data is used for training and the other half for testing. We used the multi-class Area Under the Curve (AUC) and balanced accuracy (Bacc) to measure the classifiers' performance. The AUC is calculated under one-vs-all settings and micro-averaging.

Table~\ref{table:supervised-attr} presents the results. Both RF and XGBoost perform similarly, regardless of the features used. All investigated artifacts and features yield high results, demonstrating that the features can encapsulate the fingerprint of each source data in a closed-set scenario.

\begin{table}[t]
\centering
\caption{Cross-Validation Closed-set Attribution Results}
\resizebox{\columnwidth}{!}{%
\begin{tabular}{lcc|cc}
\toprule
\multicolumn{1}{c}{\textbf{Feature}} & \multicolumn{2}{c}{\textbf{Bacc}} & \multicolumn{2}{c}{\textbf{AUC}} \\
\cmidrule{2-3} \cmidrule{4-5}
& \textbf{RF} & \textbf{XGBoost} & \textbf{RF} & \textbf{XGBoost} \\
\midrule
\textit{DINOv1 (BNE)} & 0.989 & 0.995 & 0.999 & \textbf{1.000} \\
\textit{DINOv2 (BNE)} & 0.968 & 0.985 & 0.998 & 0.999 \\
\textit{CLIP (BNE)} & 0.887 & 0.923 & 0.988 & 0.994 \\
\midrule
\textit{Mandelli et al}.~\cite{Mandelli2022} & 0.968 & 0.977 & 0.998 & 0.999 \\
\textit{GLCM (BNE)} & 0.987 & 0.991 & 0.999 & 0.999 \\
\textit{FFT-GLCM (BNE)} & 0.952 & 0.961 & 0.997 & 0.998 \\
\midrule
\textit{Synthbuster}~\cite{Bammey2024} & 0.956 & 0.970 & 0.998 & 0.999 \\
\textit{FFT-PEAKS (BNE)} & 0.988 & 0.991 & 0.999 & 0.999 \\
\textit{PATCH-FFT-PEAKS (BNE)} \ \ \ \ \ \ \ \ \  & \textbf{0.996 }& \textbf{0.996} & \textbf{1.000} & \textbf{1.000} \\
\bottomrule
%BNE: Result from the best noise extraction technique.
\end{tabular}
}
\label{table:supervised-attr}
\vspace{-0.4cm}
\end{table}

\paragraph{\textbf{Open-set Scenario}}
This task simulates a more challenging scenario without information about the generative models employed to create the synthetic images. Specifically, we train a one-class classifier using data from genuine sources and evaluate it using synthetic and pristine data. Our goal is to determine if the features of pristine data are distinguishable from those of synthetic data, which were not seen during training.

To make the scenario even more realistic and challenging, we consider two genuine Western blot sources, and we assess the model's ability to generalize between them. The first pristine source includes Western blots from~\cite{Mandelli2022}, extracted from scientific articles. The second source consists of raw Western blot data downloaded from Figshare, a scientific repository for raw data release.
Note that pristine data in~\cite{Mandelli2022} likely underwent post-processing, such as compression and contrast adjustment, typically used during articles' preparation. In contrast, the Figshare raw dataset consists of unprocessed images stored in TIFF files.

In our experiments, we adopt a cross-validation protocol with a train set of 3,000 genuine data samples from~\cite{Mandelli2022} and a test set of 3,000 samples from Figshare, along with the rest of synthetic data collected from~\cite{Mandelli2022}. Then, we swap the genuine source data in each split and repeat the experiment. This setup prevents the model from overfitting to one pristine data source. It provides a more realistic scenario where the training pristine data source may differ from the data encountered during inference.

We use Isolation Forest (IF) and Probabilistic PCA (PPCA) as one-class classifiers. We employ the scikit-learn~\cite{scikit-learn} implementations of IF and PPCA with default settings. 
PPCA's main components captured 95\% of the variance.
During the evaluation, we calculate Bacc using the likelihood threshold that maximizes this metric, aiming to find an upper bound for the artifacts and classifiers.

Table~\ref{table:openset-classification} presents the open-set results. The best-performing Bacc feature is \textit{FFT-PEAKS} with the Gaussian kernel for noise extraction. \textit{FFT-GLCM} outperforms this task's baselines and the \textit{GLCM} technique and achieves results comparable to \textit{FFT-PEAKS}.
Notably, in this scenario, deep-learning features performs lower than explicable artifacts. Both baselines are improved by exploring different residual noise techniques, as proposed in the work herein.

\begin{table}[t]
\centering
\caption{Cross-validation Open-set Classification Results}
\resizebox{\columnwidth}{!}{%
\begin{tabular}{lcc|cc}
\toprule
\multicolumn{1}{c}{\textbf{Feature}} & \multicolumn{2}{c}{\textbf{Bacc}} & \multicolumn{2}{c}{\textbf{AUC}} \\
\cmidrule{2-3} \cmidrule{4-5}
                & \textbf{IF} & \textbf{PPCA} & \textbf{IF} & \textbf{PPCA} \\
\midrule
DINOv1 (BNE)          & 0.799 & 0.831 & 0.847 & 0.880 \\
DINOv2 (BNE)          & 0.765 & 0.796 & 0.820 & 0.852 \\
CLIP (BNE)            & 0.746 & 0.780 & 0.750 & 0.848 \\
\midrule
Mandelli et al.~\cite{Mandelli2022} & 0.704 & 0.533 & 0.746 & 0.482 \\
GLCM (BNE)            & 0.792 & 0.834 & 0.856 & 0.875 \\
FFT-GLCM (BNE)        & \textbf{0.865} & 0.860 & 0.890 & 0.885 \\
\midrule
Synthbuster~\cite{Bammey2024} & 0.833 & 0.533 & 0.840 & 0.518 \\
FFT-PEAKS (BNE)       & 0.856 & \textbf{0.865} & 0.934 & \textbf{0.933} \\
PATCH-FFT-PEAKS (BNE) \ \ \ \ \ \ \ \ \ \ \ \ \ \ \ \   & 0.863 & 0.864 & \textbf{0.937} & 0.926 \\

\bottomrule
\end{tabular}%
}
\label{table:openset-classification}
\vspace{-0.5cm}
\end{table}

\paragraph{\textbf{One-vs-rest Source Attribution}}
This scenario investigates whether the artifacts from each generator can be distinguished when the classifier is trained using only one known data source, which can be pristine or synthetic. 
To this purpose, we train a one-class classifier for each of the five sources from the dataset in~\cite{Mandelli2022}. 
The query image is attributed to the model that provides the highest likelihood. 
It is worth noticing that this approach can be easily extended to an open-set configuration, as a likelihood threshold can be used to decide whether an input image belongs to a known source. For example, if classifier A has the highest likelihood \(\mathcal{L}\) among the classifiers but it is still below a confidence threshold \(t\), the input can be considered unknown.

We use a two-fold cross-validation approach where 3,000 images from each synthetic data source are included in each split. Additionally, 3,000 genuine Western blots from dataset~\cite{Mandelli2022} and 3,000 from Figshare are included in different splits, similarly to the open-set scenario.
We use the same one-class classifiers from the open-set scenario and measure their performance using AUC (micro-averaging) and Bacc.

\begin{figure}[t] 
    \centering
    \resizebox{\columnwidth}{!}{\input{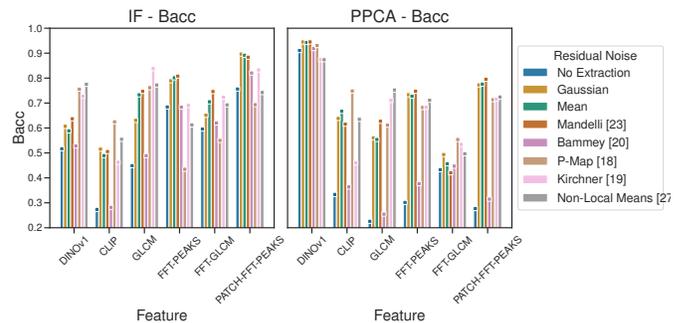}}
    \caption{One-vs-rest source attribution balanced accuracy evaluated over different residual noise extractions. Each color bar depicts a different residual noise indicated by the figure legend.}
    \label{fig:noise-exploration-attr-bacc}
    \vspace{-0.4cm}
\end{figure}

\begin{figure}[t] 
    \centering
    \resizebox{\columnwidth}{!}{\input{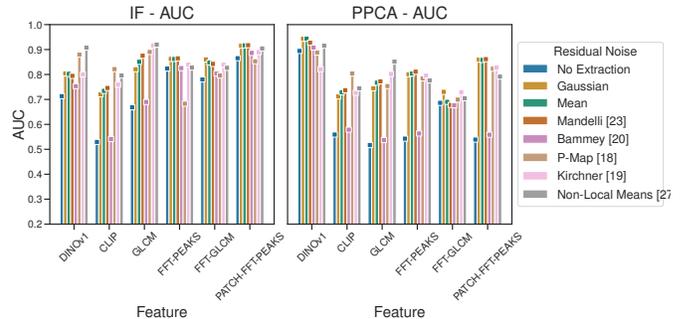}}
    \caption{One-vs-rest source attribution AUC evaluated over different residual noise extraction techniques.
    Each color bar depicts a different residual noise indicated by the figure legend.}
    \label{fig:noise-exploration-attr-auc}
\vspace{-0.4cm}
\end{figure}

Table~\ref{table:one-vs-rest-attr} presents the results for the one-vs-rest attribution task. Unlike the open-set task, DINOv1 achieves the best Bacc (0.945) and AUC (0.944) using the kernel $T$ from~\cite{Mandelli2022} during the residual noise extraction and PPCA classifier. As indicated by Fig.~\ref{fig:noise-exploration-attr-bacc} and \ref{fig:noise-exploration-attr-auc}, if no noise extraction is performed, DINOv1's Bacc drops to 0.910 and AUC to 0.894 with PPCA. 
In this scenario, \textit{PATCH-FFT-PEAKS} outperforms the baseline and \textit{FFT-PEAKS}, achieving the best Bacc among explainable methods. It also shows AUC results comparable to \textit{GLCM}.
Notably, for the IF classifier, Synthbuster improves its Bacc by 12 percentage points (pp) and Mandelli's texture-based method by 9~pp when using different residual noise techniques other than their original suggestions.

\begin{table}[t]
\centering
\caption{Cross-Validation One-vs-rest Attribution Results}
\resizebox{\columnwidth}{!}{%
\begin{tabular}{lcc|cc}
\toprule
\multicolumn{1}{c}{\textbf{Feature}} & \multicolumn{2}{c}{\textbf{Bacc}} & \multicolumn{2}{c}{\textbf{AUC}} \\
\cmidrule{2-3} \cmidrule{4-5}
                & \textbf{IF} & \textbf{PPCA} & \textbf{IF} & \textbf{PPCA} \\
\midrule
DINOV1 (BNE)         & 0.775 & \textbf{0.945} & 0.908 & \textbf{0.944} \\
DINOV2 (BNE)         & 0.702 & 0.923 & 0.865 & 0.938 \\
CLIP  (BNE)          & 0.625 & 0.748 & 0.821 & 0.804 \\
\midrule
Mandelli et al.~\cite{Mandelli2022} & 0.747 & 0.627 & 0.875 & 0.773 \\
GLCM (BNE)           & 0.838 & 0.752 & \textbf{0.920} & 0.852 \\
FFT-GLCM (BNE)        & 0.746 & 0.553 & 0.860 & 0.730 \\
\midrule
Synthbuster~\cite{Bammey2024} & 0.683 & 0.377 & 0.826 & 0.563 \\
FFT-PEAKS (BNE)       & 0.808 & 0.747 & 0.865 & 0.811 \\
PATCH-FFT-PEAKS (BNE)\ \ \ \ \ \ \ \ \ \ \ \ \ \ \ \    & \textbf{0.896} & 0.795 & 0.917 & 0.861 \\

\bottomrule
%BNE: Result from the best noise extraction technique.
\end{tabular}%
}
\label{table:one-vs-rest-attr}
\vspace{-0.3cm}
\end{table}

\paragraph{\textbf{Noise Extraction Impact}}
In this experiment, we explore the responses of the presented features and artifacts to a range of residual noise extraction techniques. We employ the methods outlined in Section~\ref{sec:exploring-residual-noise} to extract the residual noise at each artifact's workflow. We then proceed to evaluate the impact of each noise extractor by testing it on the one-vs-rest attribution task using the IF and PPCA classifiers.

Fig.~\ref{fig:noise-exploration-attr-bacc} presents the Bacc results for this setup, while Fig.~\ref{fig:noise-exploration-attr-auc} shows the AUC. We omit DINOv2 feature from this experiment since it behaves similarly to DINOv1. We include a more detailed version of these figures in the Supplementary Materials.

The importance of choosing a proper residual noise extraction technique for each feature and artifact becomes evident by analyzing the figures. For instance, features such as CLIP and \textit{GLCM} show about 40~pp variation in Bacc depending on the choice of noise extraction for the IF and PPCA classifiers.
As expected, the least favorable outcomes for most features manifest when no noise extraction is implemented (i.e., ``No Extraction''). Moreover, selecting a suitable residual noise extraction method can enhance Synthbuster and Mandelli et al.'s techniques' original implementations, underscoring our findings' relevance.

\vspace{-0.15cm}
\section{Discussion}
Our experiments aimed to understand how hand-crafted explainable artifacts and data-driven features perform in a source attribution task, avoiding black-box classifiers.

We began our research with a closed-set scenario, where all data sources were known during training. As anticipated, all artifacts demonstrated excellent results, providing a strong foundation for our study.

In a more complex scenario, with an open set formulation, we tested the artifacts using one-class classifiers trained only on genuine data. During testing, the detectors had to distinguish pristine from synthetic content. Results indicated that hand-crafted features were more effective than deep learning ones, with \textit{PATCH-FFT-PEAKS} achieving one of the best performances (0.863 Bacc and 0.937 AUC with Isolation Forest as the classifier and using kernel $T$ while extracting the residual noise). The open-set results highlighted potential biases of deep-learning features and the generalization properties of the hand-crafted ones.

Our research has practical implications, particularly in the context of real-world problems like tracking paper mills. We performed a one-vs-all attribution using one-class classifiers, a method that can be extended to an open-set scenario. Our experiments showed promising results with pre-trained deep-learning features from DINO-v1, especially with Probabilistic PCA. Combined with explainable artifacts like \textit{PATCH-FFT-PEAKS}, this approach could enhance source attribution tasks, as demonstrated by their strong performance with Isolation Forest.

We further explored the impact of different residual noise extraction techniques on the performance of each investigated feature. Our results showed that different noise extraction methods benefited each feature type (periodic-based, texture-based, or deep-learning-based). For instance, \textit{FFT-PEAKS} showed a 12~pp improvement in Bacc using kernel $T$ instead of its original method. Future work should investigate how residual noise correlates with each artifact and explore combinations of noise extraction techniques that could enhance each artifact's exposure.

\section{Conclusions and future work}
With the advancement of generative AI models, paper mills may soon use these technologies to scale their production of fake content.
By investigating the problem, we found that a possible way to track paper mills is to detect the AI models they might be using.
Our work focused on Western blots due to their 
prevalence and vulnerability to paper mills.
These images have already been synthetically generated by AI~\cite{Mandelli2022} and have proven difficult for experts to distinguish them~\cite{Qi2020}. 

As different models may accentuate different types of artifacts, our work did not aim to design a single solution that could track all AI models (which might not exist) but to develop and promote explainable solutions to expose synthetic data.

Our analysis focused on low-level artifacts from periodic and texture features left as fingerprints by the AI models. We improved state-of-the-art periodic features~\cite{Bammey2024} for source attribution by combining different image patches and analyzing their resultant Fourier spectrum. We also improved the state-of-the-art texture artifacts~\cite{Mandelli2022} in an open-set task, where models were trained only with artifacts from genuine data and tested with both synthetic and pristine data.

Another contribution of our work was exploring different types of residual noise extractors for source attribution. The experiments indicated that this part of the workflow is crucial for exposing synthetic artifacts. By choosing a better residual noise extractor, we improved Synthbuster by 12~pp and Mandelli et al.'s work by 9~pp in terms of balanced accuracy when using Isolation Forest for a one-vs-rest source attribution task.

In addition to the analysis and the proposed explainable features, our work aims to foster addressing the problem of paper mills, calling on the forensic community for more discussions, research, and solutions to this serious issue. Thus, possible paths to continue this research should consider other types of generative models and scientific images. Additionally, one should investigate new residual noise extraction techniques to expose AI fingerprints. Also,  future open-set solutions could be important for detecting unknown generative models. Finally, another promising research path is exploiting other artifacts derived from the linear combination of AI-generated pixels, linking this aspect to the generative models.

\end{document}